\documentclass[runningheads]{llncs}
\usepackage{array}
\newcolumntype{P}[1]{>{\centering\arraybackslash}p{#1}}
\usepackage{graphicx}
\usepackage{caption}
\usepackage{subcaption}
\usepackage{enumerate}
\usepackage{makecell}
\usepackage{amsmath}
\usepackage{times}
\usepackage{xcolor}
\usepackage{multirow}

\usepackage[utf8]{inputenc} 
\usepackage[T1]{fontenc}    
\usepackage{hyperref}       
\usepackage{url}            
\usepackage{booktabs}       
\usepackage{nicefrac}      
\usepackage{microtype}  
\usepackage{doi}
\usepackage{algpseudocode} 
\usepackage{algorithm}
\usepackage{adjustbox}

\begin{document}
\title{Active Inference for Stochastic Control}
%
%
\author{Aswin Paul\inst{1,2,3} \and Noor Sajid\inst{4} \and
 Manoj Gopalkrishnan\inst{2} \and
 Adeel Razi\inst{3,4,5,6}}

 \authorrunning{A. Paul et al.}

%
 \institute{IITB-Monash Research Academy, Mumbai, India  \and
 Department of Electrical Engineering, IIT Bombay, Mumbai, India \and 
 Turner Institute for Brain and Mental Health, Monash University, Australia \and
 Wellcome Trust Centre for Human Neuroimaging, UCL, United Kingdom\and
 Monash Biomedical Imaging, Monash University, Australia \and
 CIFAR Azrieli Global Scholars Program, CIFAR, Toronto, Canada 
 }
\maketitle              

\begin{abstract}
Active inference has emerged as an alternative approach to control problems given its intuitive (probabilistic) formalism. However, despite its theoretical utility, computational implementations have largely been restricted to low-dimensional, deterministic settings. This paper highlights that this is a consequence of the inability to adequately model stochastic transition dynamics, particularly when an extensive policy (i.e., action trajectory) space must be evaluated during planning. Fortunately, recent advancements propose a modified planning algorithm for finite temporal horizons. We build upon this work to assess the utility of active inference for a stochastic control setting. For this, we simulate the classic windy grid-world task with additional complexities, namely: $1)$ environment stochasticity; $2)$ learning of transition dynamics; and $3)$ partial observability. Our results demonstrate the advantage of using active inference, compared to reinforcement learning, in both deterministic and stochastic settings. 
\end{abstract}

\section{Introduction}

Active inference, a corollary of the free energy principle, is a formal way of describing the behaviour of self-organising systems that interface with the external world and maintain a consistent form over time \cite{Karl_2010,kaplan2018planning,Franz_2020}. Despite its roots in neuroscience, active inference has snowballed to many fields owing to its ambitious scope as a general theory of behaviour \cite{oliver2019active,rubin2020future,deane2020losing}. Optimal control is one such field, and several recent results place active inference as a promising optimal control algorithm \cite{Karl_2009,friston2012active,Noor_2021}. However, research in the area has largely been restricted to low-dimensional and deterministic settings where defining, and evaluating, policies (i.e., action trajectories) is feasible \cite{Noor_2021}. This follows from the active inference process theory that necessitates equipping agents a priori with sequences of actions in time. For example, with $8$ available actions and a time-horizon of $15$, the total number of (definable) policies that would need to be considered $ \to 3.5\times 10^{13}$.

This becomes more of a challenge in stochastic environments with inherently uncertain transition dynamics, and no clear way to constrain the large policy space to a smaller subspace. Happily, recent advancements like sophisticated inference \cite{Karl_2021} propose a modified planning approach for finite-temporal horizons \cite{Lance_2020}. Briefly, sophisticated inference \cite{Karl_2021}, compared to the earlier formulation \cite{Lance2020a,Noor_2021}, provides a recursive form of the expected free energy that implements a deep tree search over actions (and outcomes) in the future. We reserve further details for Section~\ref{sec:efe}.

In this paper, we evaluate the utility of active inference for stochastic control using the sophisticated planning objective. For this, we utilise the windy grid-world task \cite{Sutton1998}, and assess our agent's performance when varying levels of complexity are introduced e.g., stochastic wind, partial observability, and learning the transition dynamics. Through these numerical simulations, we demonstrate that active inference, compared to a Q-learning agent \cite{Sutton1998}, provides a promising approach for stochastic control.

\section{Stochastic control in a windy grid-world}
\begin{figure}[!ht]
\centering
\includegraphics[width=70mm,scale=1]{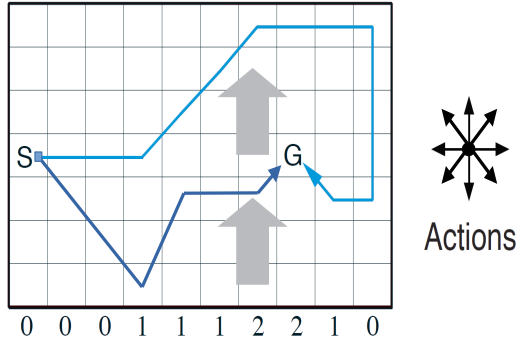}
\caption{Windy grid-world task. Here, $S$ and $G$ denote starting and goal locations. On the x-axis, the wind amplitude is shown. This is quantified as the number of unintended additional columns the agent moves during each action e.g., any action in column four results in one unintended shift upwards. There are $8$ actions: $N,S,E,W,NW,SW,SE,NE$. We plot sample paths from the start to the goal state in light and dark blue. Notice, the indirect journey to the goal is a consequence of the wind.} \label{fig1}
\end{figure}

In this section, we describe the windy grid-world task, with additional complexity, used for evaluating our active inference agent (Section~\ref{sec::activeinference}). This is a classic grid-world task from reinforcement learning \cite{Sutton1998}, with a predefined start ($S$) and goal ($G$) states (Fig.~\ref{fig1}). The aim is to navigate as optimally (i.e., within a minimum time horizon) as possible, taking into account the effect of the wind along the way. The wind runs upward through the middle of the grid, and the goal state is located in one such column. The strength of the wind is noted under each column in Fig.~\ref{fig1}, and its amplitude is quantified by the number of columns shifted upwards that were unintended by the agent. Here, the agent controls its movement through $8$ available actions (i.e., the King's moves): North ($N$), South ($S$), East ($E$), West ($W$), North-West ($NW$), South-West ($SW$), South-East ($SE$), and North-East ($NE$). Every episode terminates either at the allowed time horizon, or when the agent reaches the goal state.
 
\subsection{Grid-world complexity}
To test the performance of our active inference agent in a complex stochastic environment, we introduced different complexity levels to the windy grid-world setting (Table~\ref{table:setup}).
\vspace{-3mm}
\begin{table}[!ht]
\centering
\caption{Five complexity levels for the windy grid-world task}\label{table:setup}
\begin{tabular}{|P{2.5cm}||P{2.5cm}|P{2.5cm}|P{2.5cm}|}
\hline
 &    &  & Transition  \\
Level & Wind & Observability & Dynamics \\
\hline
1 & Deterministic & Full (MDP) & Known \\
2 & Stochastic & Full (MDP) & Known \\
3 & Deterministic & Full (MDP) & Learned \\
4 & Stochastic & Full (MDP) & Learned  \\
5 & Stochastic & Partial (POMDP) & Known
\\
\hline
\end{tabular}

\end{table}
\vspace{-3mm}
\subsubsection{Wind properties} In a deterministic setting, the amplitude of the wind remains constant. Conversely, in stochastic setting, for windy columns the effect varies by one from the mean values. We consider two settings: medium and high stochasticity. For medium stochasticity, the mean value is observed $70\%$ of the time and similarly $40\%$ of the time in the high stochastic case (Table~\ref{tab2}). The adjacent wind values are observed with remaining probabilities. Here, stochasticity is not externally introduced to the system, but it is inbuilt in the transition dynamics $\mathcal{B}$ (Section~\ref{sec::activeinference}) of the environment.

\vspace{-3mm}
\begin{table}
\centering
\caption{Stochastic nature of wind}\label{tab2}
\begin{tabular}{|P{2cm}||P{2.85cm}|P{2.8cm}|}
\hline
Level &  Wind amplitude static & Wind amplitude $\pm$ 1\\
\hline
Medium & 70$\%$ of the time & $15\%$ each for $\pm 1$ \\
High & 40$\%$ of the time & $30\%$ each for $\pm 1$ \\
\hline
\end{tabular}
\end{table}

\subsubsection{Observability}  In the fully observable setting, the agent is aware of the current state i.e., there is no ambiguity about the states of affair. We formalise this as a Markov decision processes (MDP). Whereas in the partially observable environment, the agent measures an indirect function of the associated state i.e., current observation. This is used to infer the current state of the agent. We formalise this as a partially observable MDP (POMDP).  Specific details of outcome modalities used in the task are discussed in Appendix~\ref{POMDPApp}.
\vspace{-3mm}
\subsubsection{Transition dynamics known to agent} In the known set-up, the agent is equipped with the transition probabilities beforehand. However, if these are not known, the agent begins the trials with a uninformative (uniform) priors and updates its beliefs (Eq.\ref{eqn:learning}) using random transitions. Briefly, random actions are sampled and transition dynamics updated to reflect the best explanation for the observations at hand. Here, the learned dynamics are used for planning.

\section{Active inference on finite temporal horizons}\label{sec::activeinference}

\subsection{Generative model}
The generative model is formally defined as a tuple of finite sets $(S,O,T,U,B,C,A)$:

\begin{itemize}
    \item [$\circ$] $s \in S:$ states where $S = \{1,2,3,..., 70\}$ and $s_{1}$ is a predefined (fixed) start state.
    \item [$\circ$]$o \in O:$ where $o=s$, in the fully observable setting, and in partial observability  $o=f(s)$\footnote{Here, outcomes introduce ambiguity for the agent as similar outcomes map to different (hidden) states. See Appendix~\ref{POMDPApp}, Table~\ref{tab3} for implementation details.} .
    \item [$\circ$] $T\in \mathbf{N}^{+}$, and is a finite time horizon available per episode. 
    \item [$\circ$] $a \in U:$ actions, where $U=\{ N,S,E,W,NW,SW,SE,NE \}$. 
    \item [$\circ$]$\mathcal{B}:$ encodes the transition dynamics, $P(s_{t} \vert s_{t-1}, a_{t-1},\mathcal{B})$ i.e., the probability that action $a_{t-1}$ taken at state $s_{t-1}$ at time $t-1$ results in $s_{t}$ at time $t$.
    \item [$\circ$]$\mathcal{C}:$ prior preferences over outcomes, $P(o|\mathcal{C})$. Here, $\mathcal{C}$ preference for the predefined goal-state.
    \item [$\circ$] $\mathcal{A}:$ encodes the likelihood distribution, $ P(o_{\tau} \vert s_{\tau}, \mathcal{A})$ for  the partially observable setting.
\end{itemize}

Accordingly, the agents generative model is defined as the following probability distribution:
\vspace{-2mm}
\begin{align}
    &P(o_{1:T},s_{1:T},a_{1:T-1},\mathcal{A},\mathcal{B},\mathcal{C}) = \\
    & P(\mathcal{A})P(\mathcal{B}) P(\mathcal{C}) P(s_{1}) \prod_{\tau=2}^{T} P(s_{\tau} \vert s_{\tau-1}, a_{\tau-1},\mathcal{B}) \prod_{\tau=1}^{T} P(o_{\tau} \vert s_{\tau}, \mathcal{A})
\end{align}
\vspace{-3mm}
\subsection{Full observability}
\subsubsection{Perception:}
During full observability, states can be directly accessed by agent with known or learned transition dynamics. Then the posterior estimates, $Q(s_{\tau+1} \vert a_{\tau},s_{\tau})$, can be directly calculated from $\mathcal{B}$ \cite{Lance_2020}.
\vspace{-2mm}
\begin{equation}
    Q(s_{\tau+1} \vert a_{\tau} , s_{\tau}) = P(s_{\tau+1} \vert a_{\tau} , s_{\tau}, \mathcal{B}).
\end{equation}
\vspace{-8mm}
\subsubsection{Planning:}\label{sec:efe}
In active inference, expected free-energy ($\mathcal{G}$) \cite{Noor_2021} is used for planning. For finite temporal horizons, the agent acts to minimise $\mathcal{G}$ \cite{Lance_2020}. Here, to calculate $\mathcal{G}$ we using the recursive formulation introduced in \cite{Karl_2021}. This is defined recursively as the immediate expected free energy plus the expected free energy for future actions:
\vspace{-2mm}
\begin{equation}
    \mathcal{G}(a_{\tau} \vert s_{\tau})=\mathcal{G}(a_{T-1} \vert s_{T-1}) = D_{KL} [Q(s_{T}\vert a_{T-1},s_{T-1}) \vert \vert C(s_{T})] 
\end{equation}
for $\tau=T-1$ and,
\begin{equation}
    \label{eq.rec.pl}
    \mathcal{G}(a_{\tau} \vert s_{\tau})=D_{KL} [Q(s_{\tau+1} \vert a_{\tau},s_{T-1}) \vert \vert C(s_{\tau+1})] ~ + ~  E_{Q}\Big[\mathcal{G}(\text{nextstep})\Big]
\end{equation}
for $\tau = 1, ..., T-2$.
\vspace{-3mm}
In Eq.\ref{eq.rec.pl}, the second term is calculated as,
\begin{equation}
    E_{Q}\Big[\mathcal{G}(\text{nextstep})\Big] = E_{Q(a_{\tau+1},s_{\tau+1} \vert s_{\tau},a{\tau})}[ \mathcal{G}(a_{\tau+1} \vert s_{\tau+1})].
\end{equation}

Prior preference over states are encoded such that the agent prefers to observe itself in the goal state at every time-step. $C(o=\text{goal}) = 1$, and $0$ otherwise. In the matrix form, the $i$th element of $C$, corresponds to $i$th state in $S$.

\subsubsection{Action selection:}
A distribution for action selection $Q(a_{\tau} \vert s_{\tau}) > 0 $ is defined using expected free energy such that,
\begin{equation}
    Q(a_{\tau} \vert s_{\tau}) = \sigma \left( -\mathcal{G} \left( U \vert s_{\tau} \right) \right).
\end{equation}
Here, $\sigma$ is the softmax function ensuring that components sum to one. At each time-step, actions are samples from:

\begin{equation}
    a_{t} \sim Q(a_{t} \vert s_{t}).
    \label{eqn:actdistri}
\end{equation}

\subsubsection{Learning transition dynamics:}
We learn the transition dynamics, $\mathcal{B}$, across time using conjugacy update rules \cite{friston2017active,Lance2020a,Noor_2021}:
\begin{equation}
    b_{a} = b_{a} + \sum_{\tau=2}^t \sum_{a \epsilon U} \delta_{a , a_{\tau}} Q(a) \left(s_{a,\tau} \otimes s_{a,\tau - 1} \right).
    \label{eqn:learning}
\end{equation}

Here, $b_{a} \sim Dir(b;\alpha)$ is the learned transition dynamics updated over time, $Q(a)$ is the probability of taking action $a$,  $s_{a,{\tau}}$ is the state at time $\tau$ as a consequence of action $a$, $s_{a,\tau - 1}$ is the state-vector at time $\tau-1$ taking action $a$, and $\otimes$ is the Kronecker-product of the corresponding state-vectors. Furthermore, we also assessed the model accuracy obtained after a given number of trials to update $\mathcal{B}$, when random actions were employed to explore transition dynamics. These learned transitions were used for control in Level-3 and Level-4 of the problem.

\subsection{Partial observability}
We formalise partial observability as a partially observed MDP (POMDP). Here, the agents have access to indirect observations about the environment. Specific details of outcome modalities used in this work are discussed in Appendix~\ref{POMDPApp}.
These outcome modalities are same for many states for e.g., the states $2$ and $11$ have the same outcome modalities (see Appendix~\ref{POMDPApp}, Table~\ref{tab3}). Here, we evaluate the ability of active inference agent to perform optimal inference and planning in the face of ambiguity. The critical advancement with sophisticated inference \cite{Karl_2021} compared to the classical formulation \cite{Noor_2021} allows us to perform deep-tree search for actions in the future. The agent infers the hidden-states by minimising a functional of its predictive distribution (generative model) of the environment called the variational free-energy. This predictive distribution can be defined as,
\begin{equation}
    Q(\Vec{s} \vert \vec{a}, \Tilde{o}) := \prod_{\tau=1}^{T} Q(s_{\tau} \vert a_{\tau-1}, s_{\tau-1} , \tilde{o}).
\end{equation}

To infer hidden-states from partial observations, thr agent engages in minimising variational free energy ($\mathcal{F}$) functional of $Q$ using variational (Bayesian) inference. For a rigorous treatment  of it, please refer to \cite{Karl_2021,Lance_2020}. In this scheme, actions are considered as random variables at each time-step, assuming successive actions are conditionally independent. This comes with a cost of having to consider many action sequences in time. The search for policies in time is optimised both by restricting the search over future outcomes which has a non-trivial posterior probability (Eg: $> 1/16$) as well as only evaluating policies with significant prior probabilities (Eg: $> 1/16$) calculated from the expected free energy (i.e., Occam's window). In the partially observable setting, the expected free energy accommodates ambiguity in future observations prioritising both preference seeking as well as ambiguity reduction in observations \cite{Karl_2021}.

\section{Results}
We compare the performance of our active inference agent with a popular reinforcement learning algorithm, Q-learning \cite{Sutton1998}, in Level $1$. Q-Learning is a model-free RL algorithm that operates by learning the 'value' of actions at a particular state. It is well suited for problems with stochastic transitions and reward dynamics due to its model-free parameterisation. Q-Learning agents are extensively used in similar problem settings and exhibit state-of-the-art (SOTA) performances \cite{Sutton1998}. To train the Q-learning agents, we used an exploration rate of $0.1$, learning rate of $0.5$ and discount factor of $1$. Training was conducted using 10 different random seeds to ensure unbiased results. The training depth for Q-Learning agents were increased with complexity of the environment.

We instantiate two Q-learning agents, one trained for $500$ time-steps (QLearning500) and another for $5000$ time-steps (QLearning5K) in Level-1. 
Both the active inference agent and the QLearning5K agent demonstrate optimal success rate for the time-horizon $T=8+$ (see Appendix \ref{results1and3}, Fig.\ref{resl1}).

\begin{figure}
\begin{subfigure}{.5\textwidth}
\centering
\includegraphics[width=2.3in]{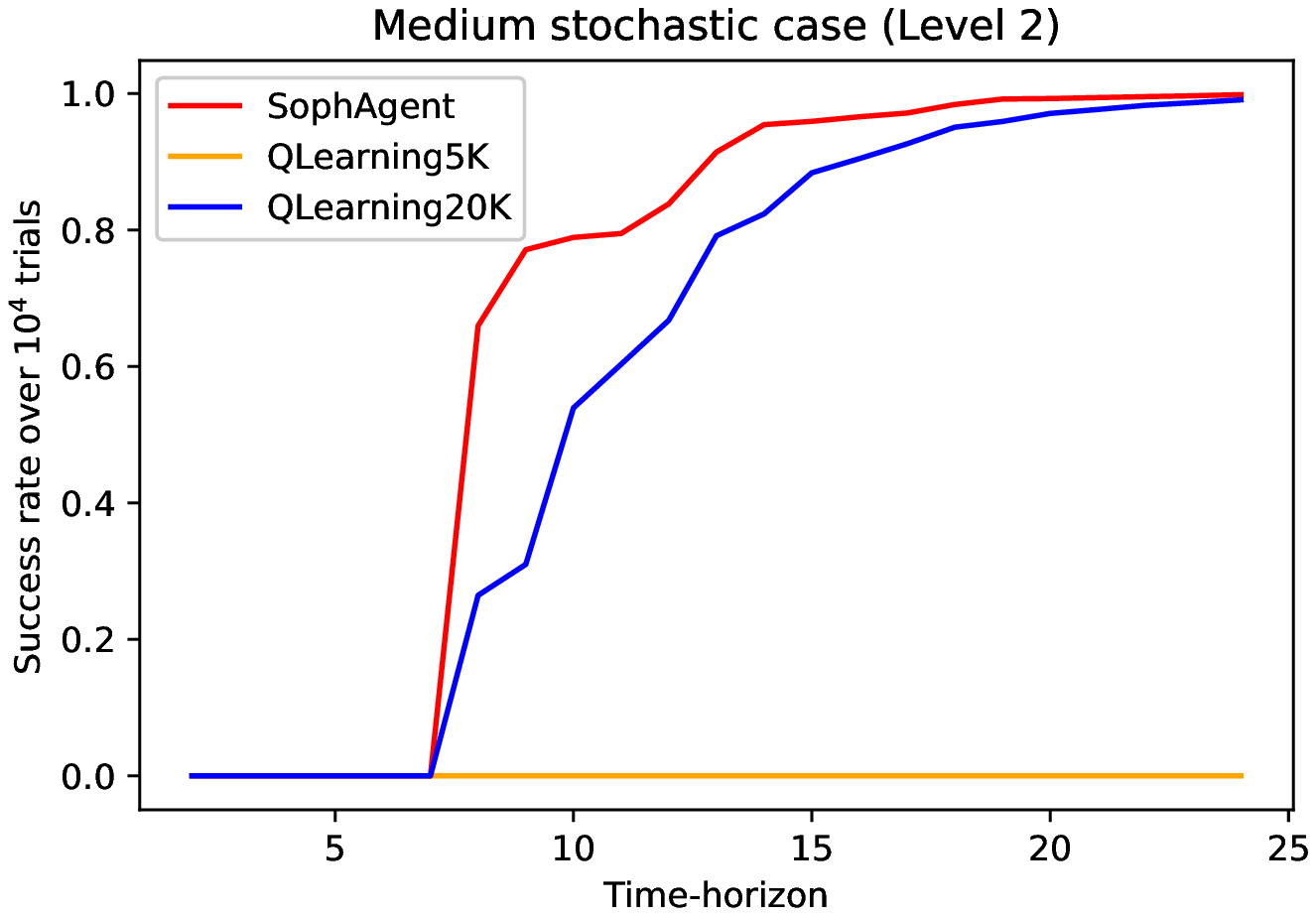}
\end{subfigure}
\begin{subfigure}{.5\textwidth}
\centering
\includegraphics[width=2.3in]{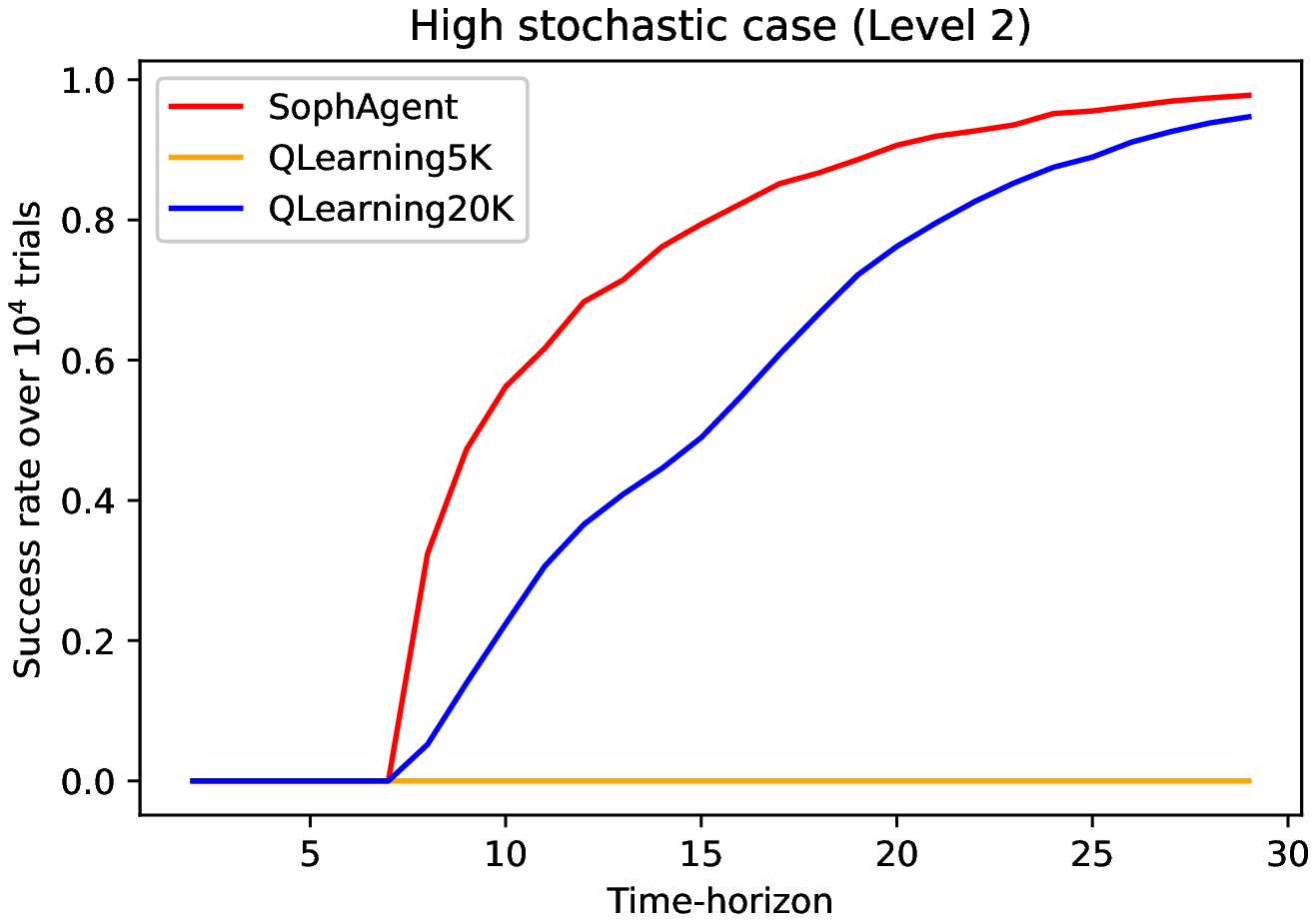}
\end{subfigure}
\begin{subfigure}{.5\textwidth}
\centering
\includegraphics[width=2.3in]{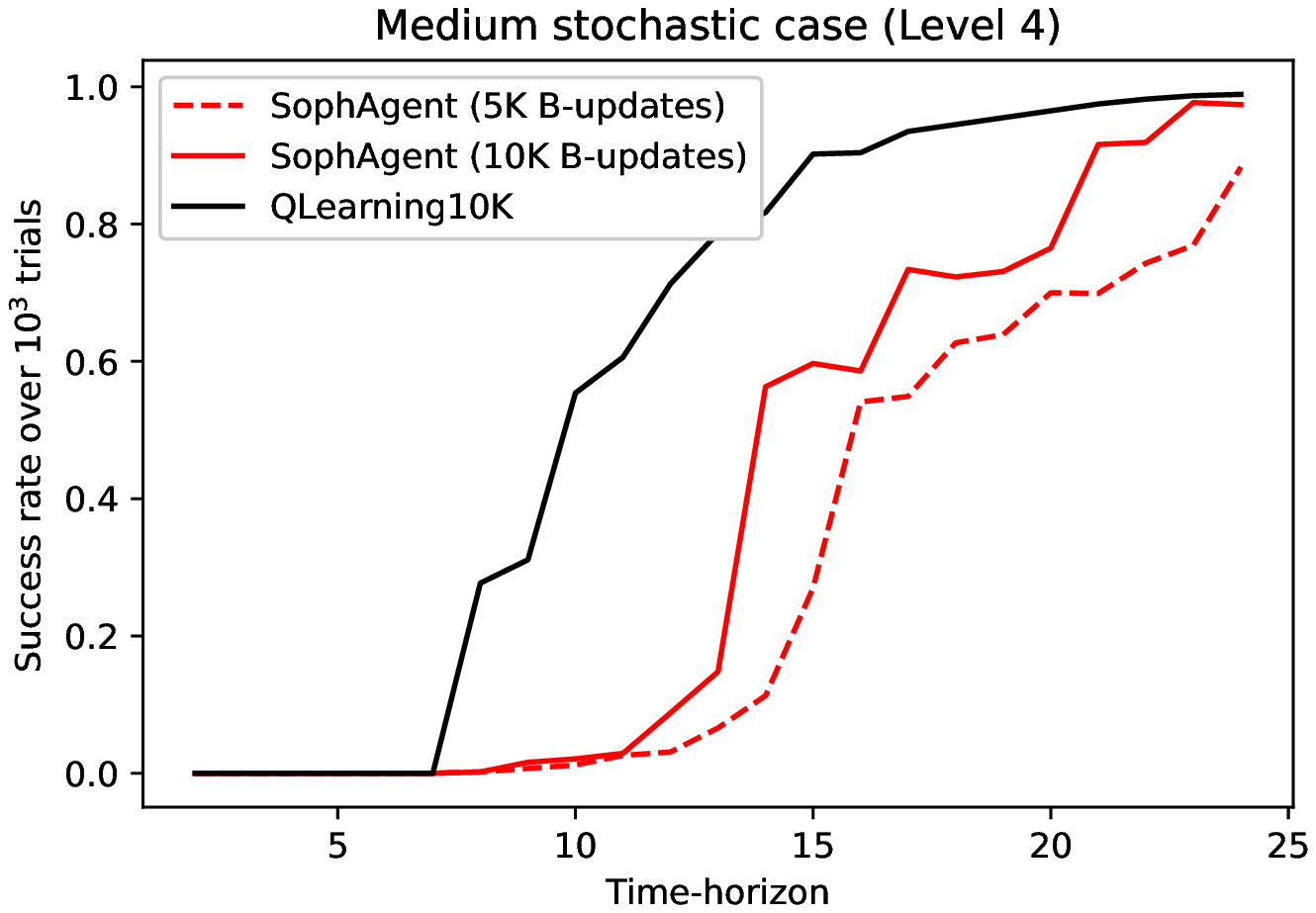}
\end{subfigure}
\begin{subfigure}{.5\textwidth}
\centering
\includegraphics[width=2.3in]{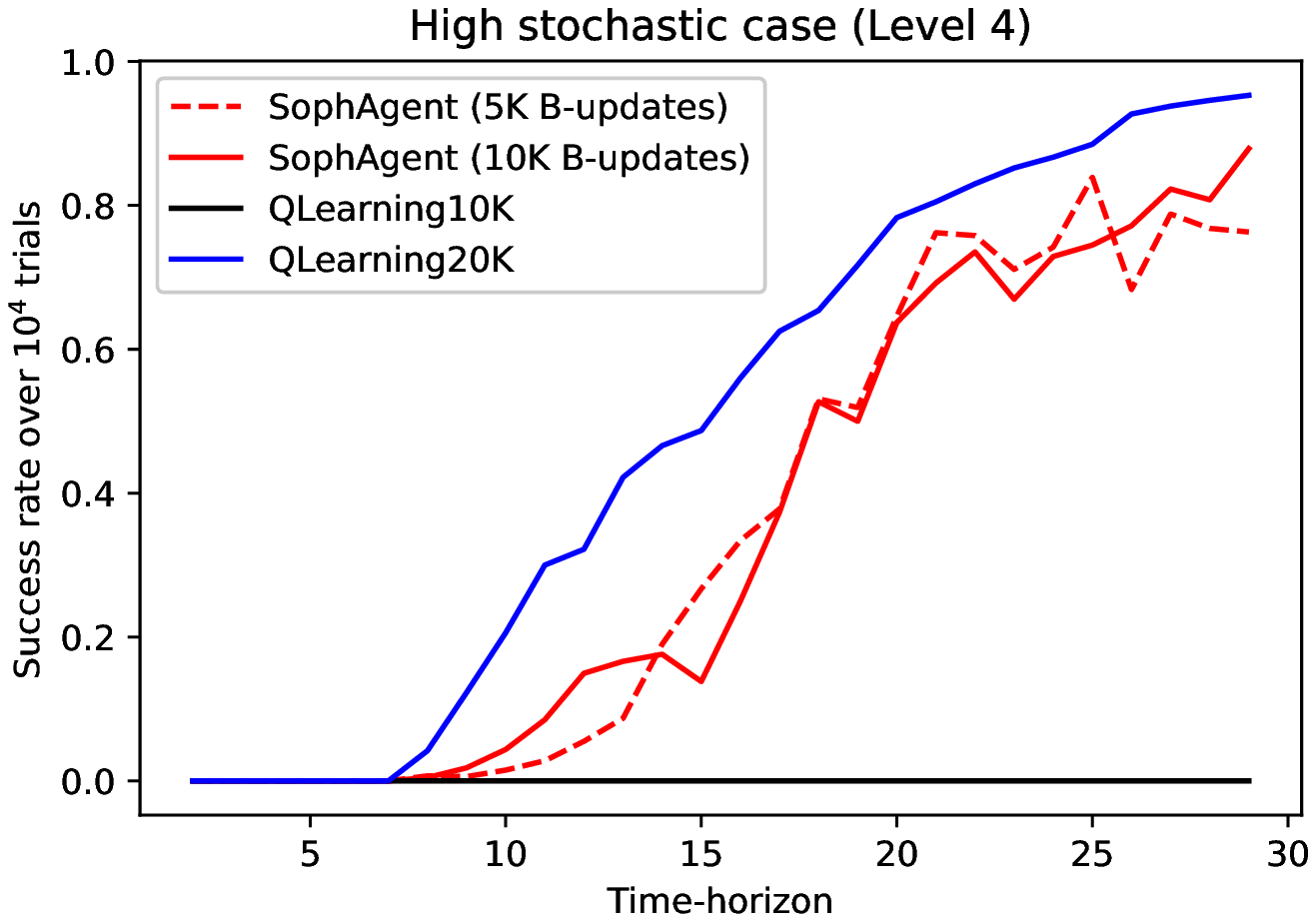}
\end{subfigure}
\begin{subfigure}{.5\textwidth}
\centering
\includegraphics[width=2.3in]{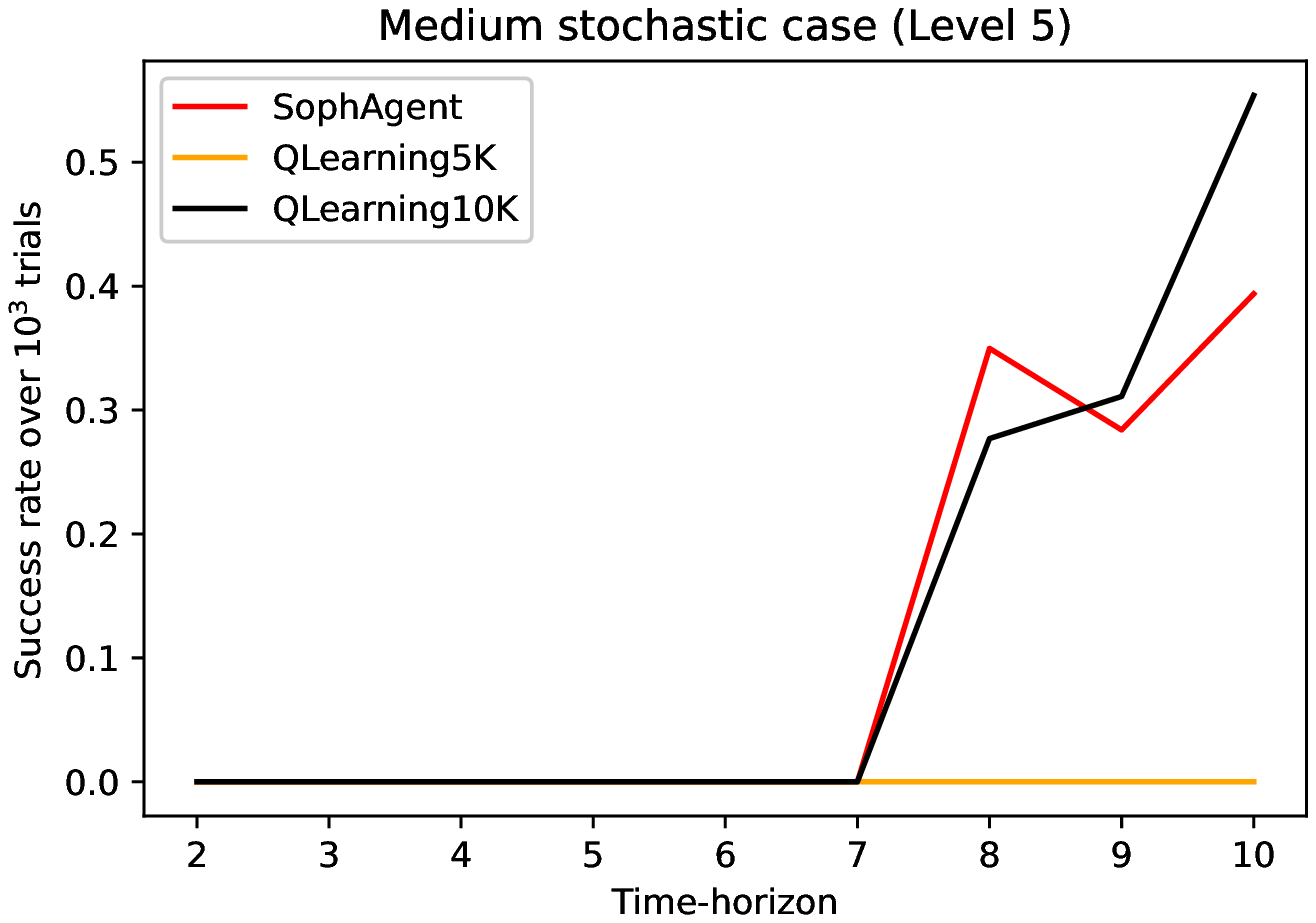}
\end{subfigure}
\begin{subfigure}{.5\textwidth}
\centering
\includegraphics[width=2.3in]{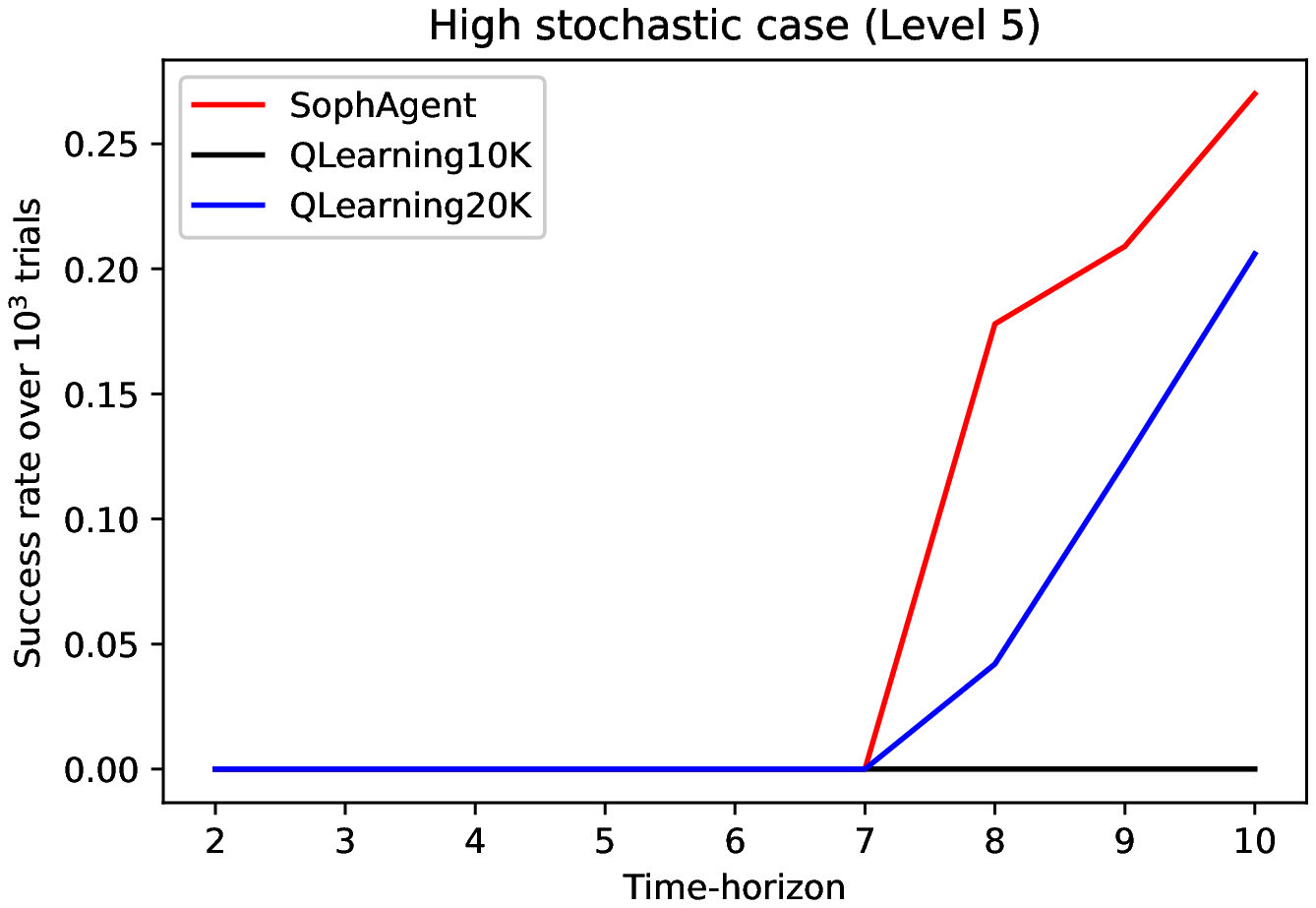}
\end{subfigure}
\caption{Stochastic environments: Performance comparison of agents in Level-2 (top row), Level-4 (middle row), and Level-5 (last row) of windy grid-world task for medium-stochastic (left column) and high-stochastic (right column) environments, respectively. Here, x-axis denotes time horizon and y-axis the success rate over multiple-trials. 'SophAgent' represents the active inference agent, 'QLearning5K' represents Q-learning agent trained for $5,000$ time-steps, 'QLearning10K' for the 'Q-learning agent trained for $10,000$ time-steps, and 'QLearning20K' for the Q-learning agent trained for $20,000$ time-steps. Each agent was trained using $10$ different random seeds. 'SophAgent (5K B-updates)' and SophAgent (10K B-updates) refers to active inference agent using self-learned transition dynamics $\mathcal{B}$ with $5000$ and $10000$ updates respectively.} 
\label{resl245}
\end{figure}

Using these baselines from the deterministic environment with known transition dynamics, we compared the performance of the agent in a complex setting with medium and highly stochastic wind (Level $2$; Table.~\ref{tab2}).Here, the active inference agent is clearly superior against the Q-Learning agents (Fig.~\ref{resl245} top row). Moreover, they demonstrate better success rates for shorter time-horizons, and 'optimal' action selection. Note, success rate is the percentage of trials for which the agent successfully reached the goal within the allowed time-horizon.

\begin{figure}
\label{modelaccuracy}
\includegraphics[width=2.3in]{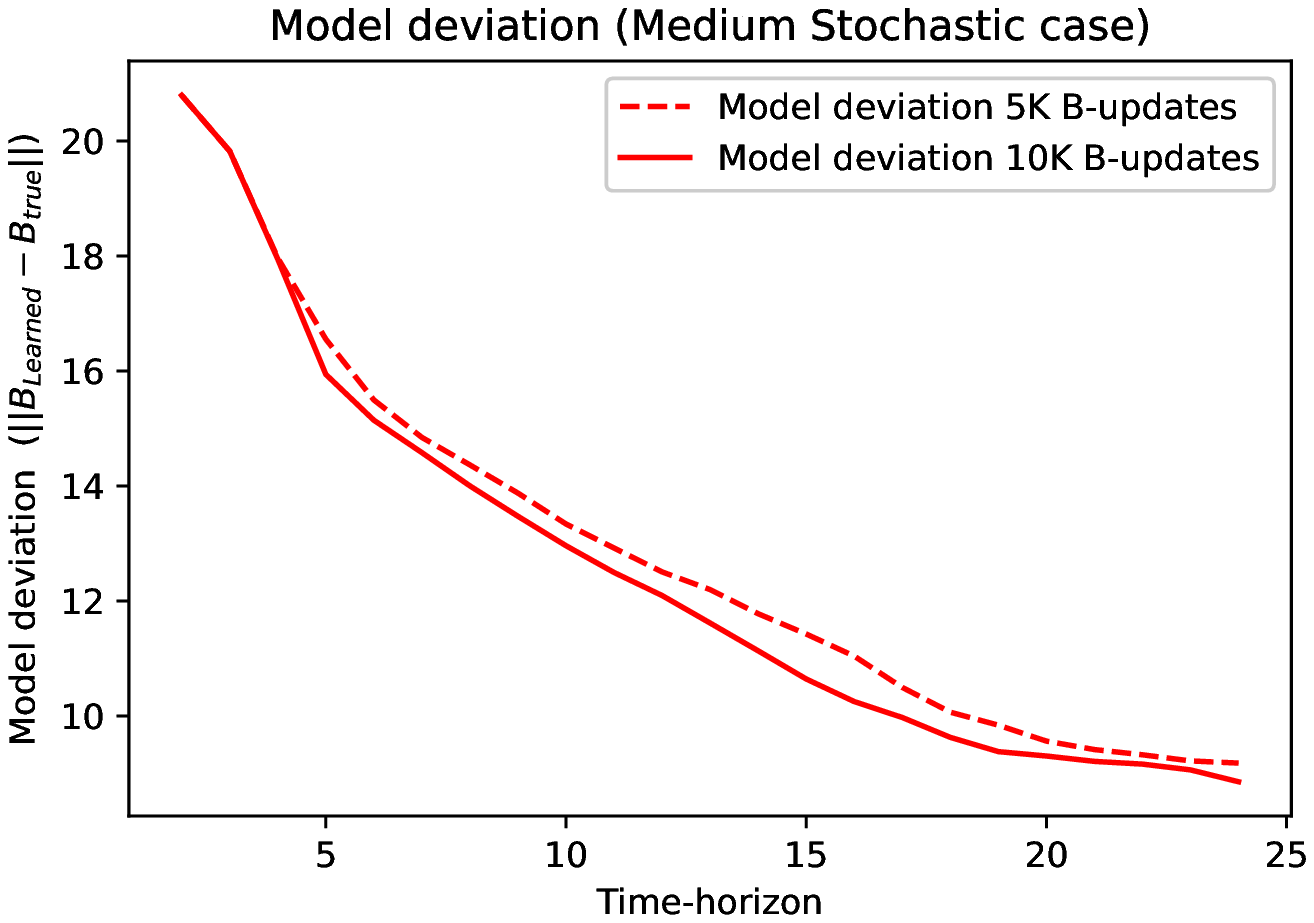}
\includegraphics[width=2.3in]{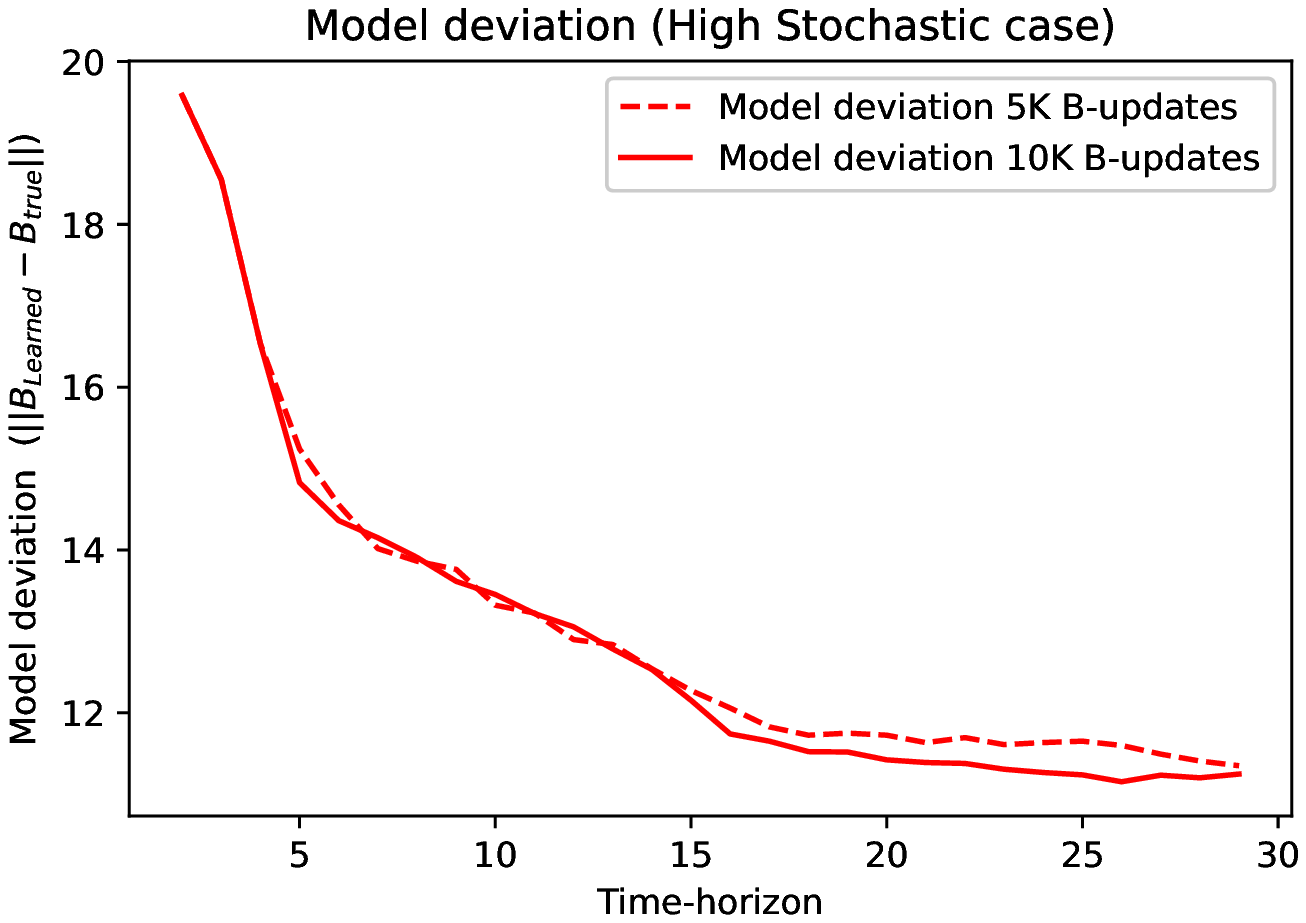}
\caption{Accuracy of learned dynamics in terms of deviation from true transition dynamics in Level-4 A: Medium stochastic case B: High stochastic case} \label{resl4b}
\end{figure}

Next, we considered how learning the transition dynamics impacted agent behaviour (Level $3$ and $4$). Here, we used Eq.~\ref{eqn:learning} for learning the transition dynamics, $\mathcal{B}$. First, the algorithm learnt the dynamics by taking random actions over $X$ steps (for example, $X$ is $5000$ time steps in 'SophAgent (5K B-updates)', see Fig. \ref{resl245} middle row). These learned transition dynamics $\mathcal{B}$ were used (see Fig.~\ref{resl4b}) by the active inference agent to estimate the action distribution in Eq.~\ref{eqn:actdistri}. Results for level $3$ are presented in Appendix~\ref{results1and3}, Fig.~\ref{resl3}. Here, the Q-Learning algorithm with $5,000$ learning steps shows superior performance to the active inference agents. However with longer time horizons, the active inference agent shows competitive performance. Importantly, the active inference agent used self-learned, and imprecise transition dynamics $\mathcal{B}$ in these levels. Level $4$ results for medium and highly stochastic setting are presented in Fig.~\ref{resl245} (middle row). For medium stochasticity, the QLearning10K exhibited satisfactory performance, however it failed with zero success rate in the highly stochastic case. This shows the need for extensive training for algorithms like Q-Learning in highly stochastic environments. However, the active inference agent demonstrated at-par performance. Remarkably, the performance was achieved using imprecise (compared to true-model), self-learned transition dynamics ($\mathcal{B}$) (see Fig.~\ref{resl4b}).

The active inference agent shows superior performance in the highly stochastic environment even with partial observability (Fig.~\ref{resl245}, last row). Conversely, excessive training was required for the Q-Learning agent to achieve a high success rate in a medium stochastic environment, but even this training depth led to a zero success rate with high stochasticity. These results present active inference, with a recursively calculated free-energy, as a promising algorithm for stochastic control.

\section{Discussion}
We explored the utility of the active inference with planning in finite temporal-horizons for five complexity levels of the windy grid-world task. Active inference agents performed at-par, or superior, when compared with well-trained Q-Learning agents. Importantly, in the highly stochastic environments the active inference agent showed clear superiority over the Q-Learning agents. The higher success rates at lower time horizons demonstrated the 'optimality' of actions in stochastic environments presented to the agent. Additionally, this performance is obtained with no specifications of acceptable policies. The total number of acceptable policies scale exponentially with the number of available actions and time-horizon. Moreover, the Level $4~ \& ~5$ results demonstrate the need for extensive training for the Q-Learning agents when operating in stochastic environments. We also demonstrated the ability of the active inference agents to achieve high success rate even with self-learned, but sub-optimal, transition dynamics. Methods to equip the agent to learn both transition-dynamics $\mathcal{B}$ and outcome-dynamics $\mathcal{A}$ for a partially observable setting have been previously explored \cite{friston2017active,Noor_2021}. For a stochastic setting, we leave their implementation for future work. 

The limitation yet to be addressed is the time consumed for trials in active inference. Large run-time restricted analysis for longer time horizons in Level $5$. Deep learning approaches using tree searches, for representing policies were proposed recently \cite{fountas2020deep,ccatal2019bayesian,deepaipomdps}, may be useful in this setting. We leave run-time analysis and optimisation for more ambitious environments for future work. Also, comparing active inference to model based RL algorithms like Dyna-Q \cite{Sutton1998} and control as inference approaches \cite{caiandai} is a promising direction to pursue. \\
We conclude that the above results place active inference as a promising algorithm for stochastic-control.

\subsubsection*{Software note}
The environments and agents were custom written in Python for fully observable settings. The script 'SPM\_MDP\_VB\_XX.m' available in SPM12 package was used in the partially observable setting. All scripts are available in the following link: \scriptsize{\url{https://github.com/aswinpaul/iwai2021_aisc}}.

\subsubsection*{Acknowledgments}
AP acknowledges research sponsorship from IITB-Monash Research Academy, Mumbai and Department of Biotechnology, Government of India. AR is funded by the Australian Research Council (Refs: DE170100128 \& DP200100757) and Australian National Health and Medical Research Council Investigator Grant (Ref: 1194910). AR is a CIFAR Azrieli Global Scholar in the Brain, Mind \& Consciousness Program. AR and NS are affiliated with The Wellcome Centre for Human Neuroimaging supported by core funding from Wellcome [203147/Z/16/Z]. \\

%
%
%
\bibliographystyle{splncs04}
%

\newpage
\begin{appendix}

{\Large\bfseries Supplementary information}

\setcounter{figure}{0}
\renewcommand*\thefigure{\Alph{section}.\arabic{figure}}

\setcounter{table}{0}
\renewcommand*\thetable{\Alph{section}.\arabic{table}}

\section{Results Level-1 and Level-3 (Non-stochastic settings)}
\label{results1and3}

\begin{figure}
\includegraphics[width=\textwidth]{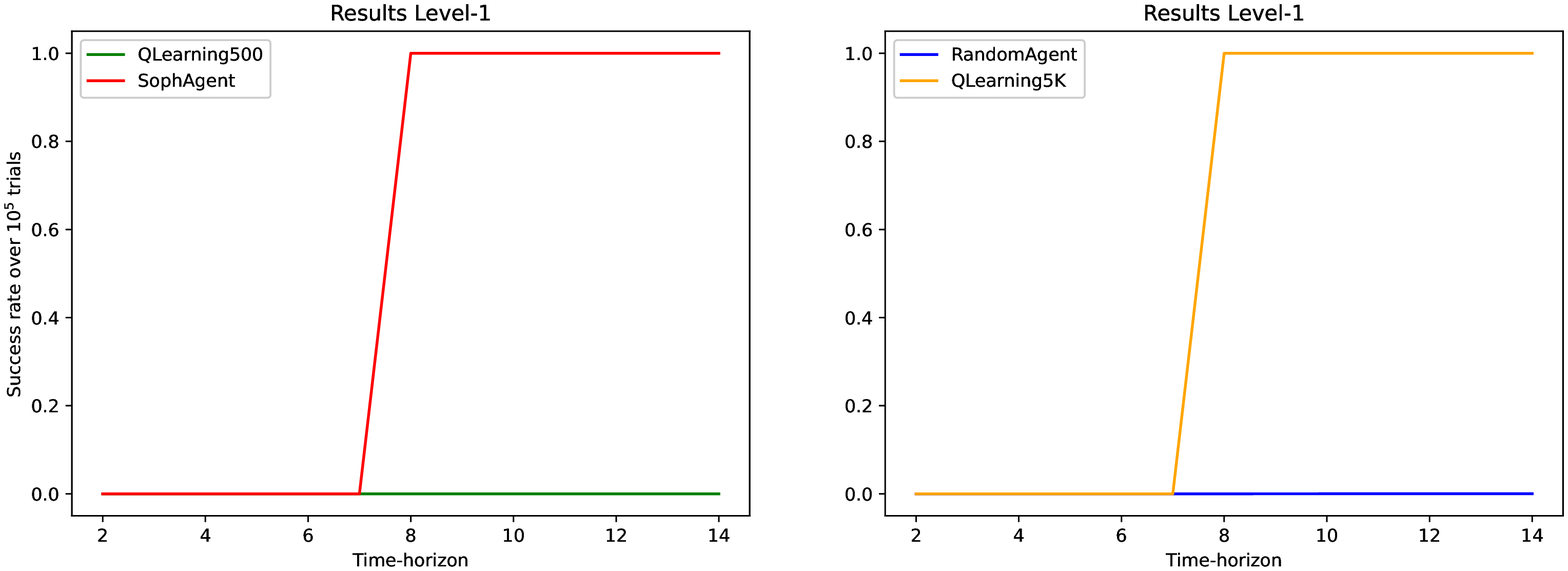}
\caption{Performance comparison of agents in Level-1 of windy grid-world task. 'RandomAgent' refers to a naive-agent that takes all actions with equal probability at every time step.} \label{resl1}
\end{figure}

\begin{figure}
\includegraphics[width=2.3in]{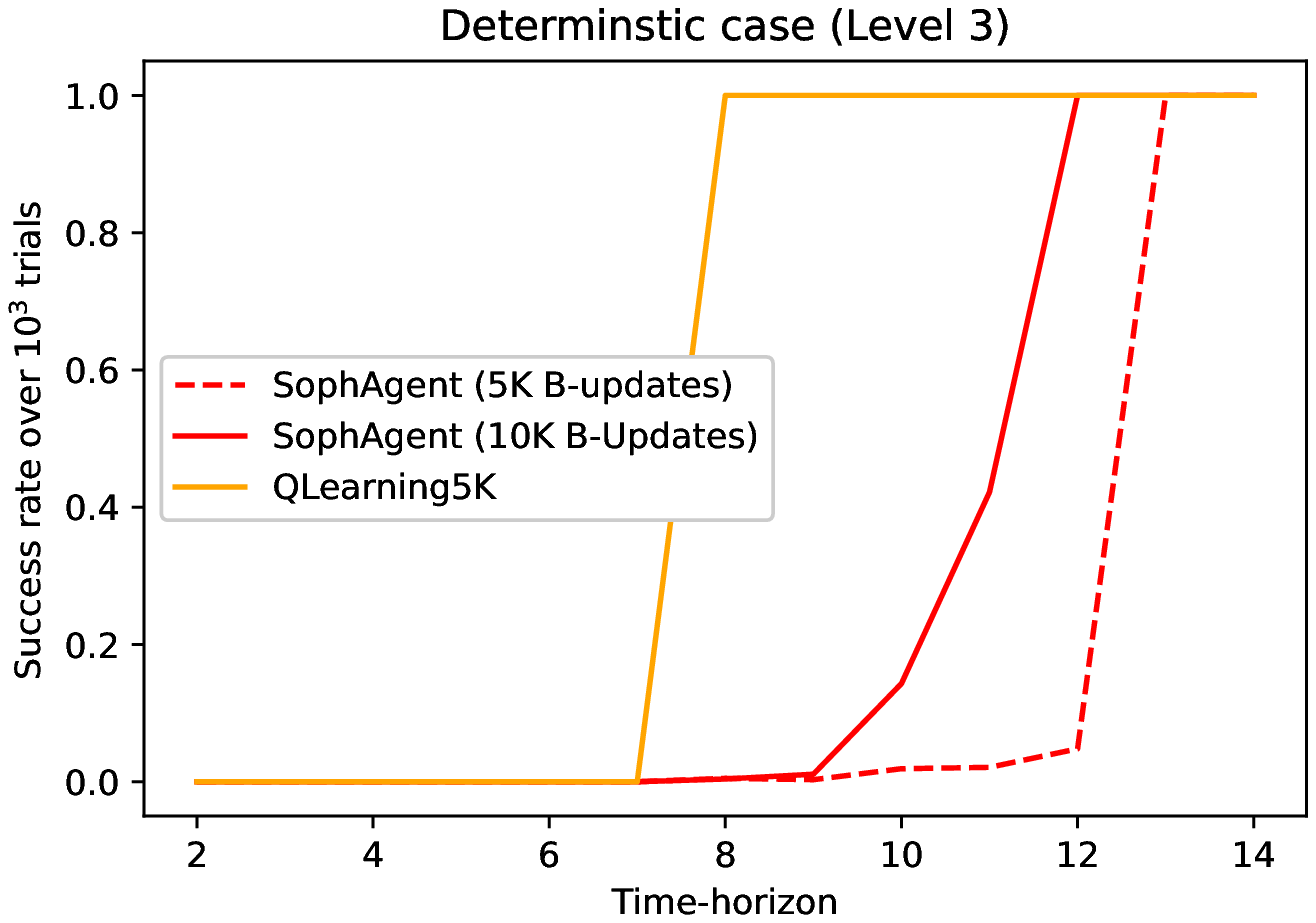} 
\includegraphics[width=2.3in]{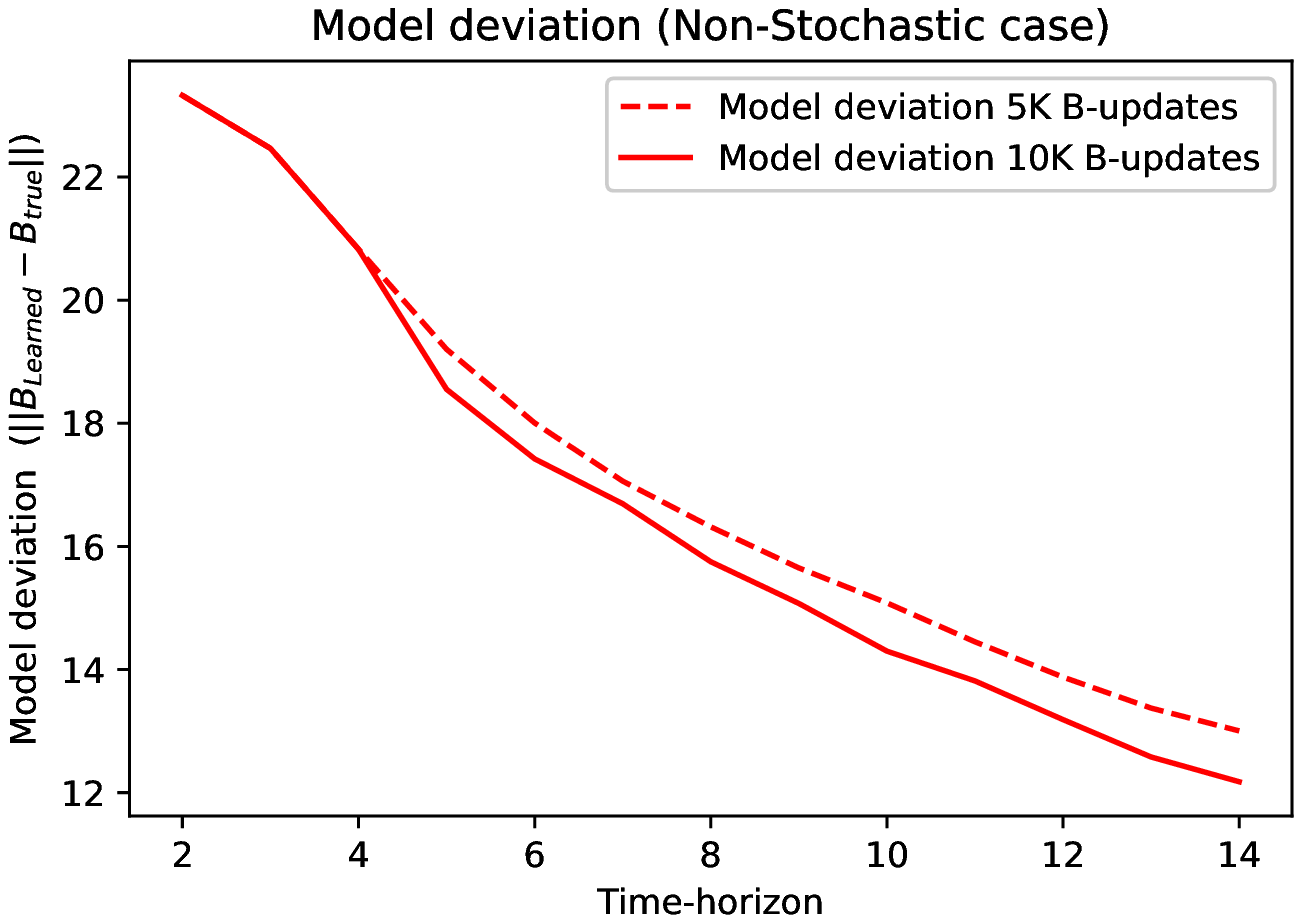}
\caption{A: Performance comparison of active inference agents with learned $B$ using 5000 and 10000 updates respectively to Q-Learning agent in Level-3. 'Q-Learning5K' stands for Q-Learning agent trained for $5000$ time steps using $10$ different random seeds. B: Accuracy of learned dynamics in terms of deviation from true dynamics.} \label{resl3}
\end{figure}

\newpage
\section{Outcome modalities for POMDPs}
\label{POMDPApp}
\normalsize{In the partially observable setting, we considered two outcome modalities and both of them were the function of 'side' and 'down' coordinates defined for every state in Fig. \ref{fig1}. Examples of the coordinates and modalities are given below. First outcome modality is the sum of co-ordinates and second modality is the product of coordinates.}

\begin{table}
\centering
\caption{Outcome modalities specifications}\label{tab3}
\begin{tabular}{|c|c|c|c|c|}
\hline
State & \makecell{Down \\ coordinate (C1)} & \makecell{Side \\ coordinate (C2)} & \makecell{Outcome-1 \\ (C1+C2)} & \makecell{Outcome-2 \\ (C1*C2)} \\
\hline
1 & 1 & 1 & 2 & 1 \\
2 & 1 & 2 & 3 & 2 \\
. & . & . & . & . \\
11 & 2 & 1 & 3 & 2 \\
. & . & . & . & . \\
31 & 4 & 1 & 5 & 4 \\
38 & 4 & 8 & 12 & 32 \\
. & . & . & . & . \\
\hline
\end{tabular}
\end{table}

These outcome modalities are similar for many states (for e.g., states $2$ and $11$ have the same outcome modalities (see Tab. \ref{tab3})). The results demonstrates the ability of active inference agent to perform optimal inference and planning in the face of ambiguity. One of the output from 'SPM\_MDP\_VB\_XX.m' is 'MDP.P'. 'MDP.P' returns the action probabilities an agent will use for a given POMDP as input at each time-step. This distribution was used to conduct multiple trails to evaluate success rate of the active inference agent.

\end{appendix}
\end{document}